\documentclass[letterpaper]{article} 
\usepackage{aaai2026}  
\usepackage{times}  
\usepackage{helvet}  
\usepackage{courier}  
\usepackage[hyphens]{url}  
\usepackage{graphicx} 
\usepackage{natbib}  
\usepackage{caption} 
\usepackage{algorithm}
\usepackage{algorithmic}

\usepackage{enumitem}
\usepackage{xcolor}
\usepackage{booktabs}
\usepackage{tabularx}
\usepackage{tikz}
\usetikzlibrary{arrows.meta}
\usepackage{amsmath}

\usepackage{newfloat}
\usepackage{listings}
\DeclareCaptionStyle{ruled}{labelfont=normalfont,labelsep=colon,strut=off} 
\floatstyle{ruled}
\newfloat{listing}{tb}{lst}{}
\floatname{listing}{Listing}
\title{Explainable AI for the EU Right to Explanation:\\A Systematic Review of the Law-XAI Translation Gap}

\author {
    Benjamin Fresz \textsuperscript{\rm 1,2},
    Elena Dubovitskaya\textsuperscript{\rm 3},
    Marco F. Huber\textsuperscript{\rm 1,2}
}
\affiliations {
    \textsuperscript{\rm 1}Fraunhofer Institute for Manufacturing Engineering and Automation IPA, Stuttgart, Germany\\
    \textsuperscript{\rm 2}Institute of Industrial Manufacturing and Management IFF, University of Stuttgart, Germany\\
    \textsuperscript{\rm 3}University of Giessen, Germany\\
    benjamin.fresz@ipa.fraunhofer.de, ORCID: 0009-0002-7463-8907,
    elena.dubovitskaya@recht.uni-giessen.de, \\ ORCID: 0000-0002-0491-3670, 
    marco.huber@ieee.org, ORCID: 0000-0002-8250-2092
}

\usepackage{bibentry}

\begin{document}

\maketitle

\begin{abstract}
When algorithms make or influence consequential decisions---about loan eligibility, hiring, or healthcare---EU law grants affected individuals a Right to Explanation.
Yet whether (and how) Explainable AI (XAI) can satisfy this right in practice remains poorly understood, with direct implications for individuals' ability to contest automated decisions that affect their lives.
This paper presents a systematic literature review of XAI in the context of the EU Right to Explanation, with particular focus on Art. 15(1)(h) GDPR, Art. 86 AI Act (AIA), and related instruments.
We consider papers published from 2024 onwards, as the final version of the AIA was published in July 2024---with Art.~86 being added late.
From 2643 initial records identified by a deliberately broad search, we review 57 full texts, of which only 19 papers demonstrate substantive integration of both legal and technical perspectives, showing gaps in the interdisciplinary synthesis of the current regulatory framework.
We document three problematic patterns across the corpus: Most misidentify the GDPR legal basis; few engage with the CJEU's Dun \& Bradstreet judgment (likely due to publication timing); and the distinction between explanation form (governed by addressee) and content (governed by legal purpose) is often conflated.
We conceptualize this as the Addressee/Purpose Framework, propose a four-phase blueprint for operationalization, and identify six concrete open research questions.
Without further progress, the Right to Explanation risks remaining a formal obligation without a technically realizable path to compliance.
\end{abstract}


\section{Introduction}
\label{sec:introduction}

The explanation of decisions has become a part of the governance of automated decision-making, given its direct implications for individual rights and obligations, although its exact relevance and necessity are still a point of discussion.
This paper describes the legal grounding of a Right to Explanation, which is now explicitly demanded in Art.\ 86 of the Artificial Intelligence Act (AIA) of the European Union. As such, relevant laws are described---namely the General Data Protection Regulation (GDPR), the AIA and the Consumer Credit Directive (CCD).

Within the context of Artificial Intelligence (AI) or Machine Learning (ML), researchers in the field of explainable AI (XAI) develop methods to ``explain'' AI decisions or models, i.e., provide additional information on how an AI model works in general (global explainability) or how a specific decision was made (local explainability). The field is subject to many influences, as there is a long history of legal and philosophical discussion on explanations \cite{Hempel1948}, social science research on learning and explanations \cite{Miller2017}, and interface design for presenting information accessibly. Within XAI, different sub-fields and terminology exist, especially the naming of explanations (as in XAI) or interpretations (as in interpretable Machine Learning, iML); this distinction is explained in Section~\ref{sec:xai-classification}.
Despite progress, XAI methods remain error-prone, i.e., explanations do not conform to the AI model in question, are difficult to evaluate objectively and for laypeople to understand, and may even be misleading \cite{Longo2024}.

From a legal perspective, there are various fields of application for XAI. The first of them is the Right to Explanation, now anchored in several EU legal instruments, with the GDPR (Art.\ 15(1)(h)) still being one of the most important. A similar right can be found in the 2023 CCD (Art.\ 18(8)(a)) recast. In 2024, the Right to Explanation in Art.\ 86 AIA was added. The Right to Explanation is a central focus of this research paper, but it is important to keep in mind that, while the Right to Explanation is the most visible application of XAI in law, it is not the only one.
Further legal grounds to use XAI may include transparency requirements (Art. 13 AIA), human oversight (Art. 14 AIA) \cite{Sapienza}, safeguarding of rule-of-law principles when AI is used in the judiciary and public administration, product liability and company law~\cite{Fresz2024}.

In recent years, multiple interdisciplinary publications have tried to bridge the gap from legal norms to implementation-ready XAI recommendations, with the earliest contributions debating whether the Right to Explanation exists in the GDPR and proposing counterfactual explanations as the most suitable approach \cite{Wachter2018}.
Since the Dun~\&~Bradstreet judgment of the Court of Justice of the European Union (CJEU, Section~\ref{sec:right-to-explanation}) settled the existence of the Right to Explanation, the debate has shifted to implementation---particularly with Art.~86 AIA and new scholarly publications.
But interdisciplinary work, especially within fields such as technology and law, which work differently and use their own sources of knowledge, remains difficult.
We provide an overview and points of critique of the state of discussion on the
implementation of the Right to Explanation via XAI.
Our core contributions include:

\begin{itemize}
    \item A clear, norm-dogmatic derivation of the Right to Explanation in
    Art.\ 15(1)(h) GDPR, Art.\ 86 AIA, and Art.\ 18(8)(a) CCD, resolving widespread
    and consequential confusion over its legal basis.

    \item Evidence of the limited integration of legal and technical insights in the post-AIA scientific literature: Of 2643 surveyed records, only 57 records meet the requirements for full text screening, of which 19 papers demonstrate meaningful engagement with both domains---itself a significant finding about the field's coverage of the current regulatory framework.

    \item Documentation of problematic patterns and omissions in the literature, including imprecise grounding of the GDPR Right to Explanation, unrealistic expectations that laws provide technical detail, and---likely an artifact of publication timing---the omission of recent case law.
    \item Conceptualization of an addressee/purpose distinction that the literature
    often conflates: The \emph{addressee} of an explanation determines its
    required \emph{form}, the \emph{purpose} determines its \emph{content}.
    This distinction provides a shared basis for discussion and concrete design guidance for practitioners developing legally compliant XAI systems.

    \item A four-phase blueprint for operationalization that describes the structured process from identifying applicable legal requirements to documenting compliance, and makes explicit where each of the six open research questions currently blocks progress.
\end{itemize}



\section{The Right to Explanation}
\label{sec:right-to-explanation}

The Right to Explanation is now enshrined in several EU legal instruments, with the GDPR still being one of the most important.
In the Dun \& Bradstreet judgment, the CJEU gave the GDPR-based Right to Explanation a much clearer shape and resolved several contentious
issues~\cite{CJEU2025}.
The Court's relevant reasoning can be summarized as follows: (1) Art.\ 15(1)(h) GDPR grants the data subject a genuine Right to Explanation of the specific automated decision, even where the processing operations are highly complex; (2) the explanation must be meaningful, i.e., useful, relevant, important and easily understandable, and must meet a
certain qualitative standard; (3) according to Art.\ 12(1) GDPR, the explanation must be provided in a form that is concise, transparent, intelligible and easily accessible, using clear and plain language; (4) merely disclosing the algorithm or providing a detailed description of every step in an automated decision-making process would not be sufficiently precise or intelligible.

The CJEU did not lay down detailed requirements regarding the substance of explanations, except that it may be sufficient to inform the data subject to what extent a variation in the personal data used would have led to a different result.
This likely refers to counterfactual explanations \cite{Wachter2018,Palazzo2025}.

At the same time, the CJEU emphasizes that explanations should, insofar as possible, be designed not to infringe the protection of trade secrets (Recitals 4 and 63 of the GDPR).
The controller therefore has the option of first transmitting the relevant information to the competent supervisory authority or the competent court, which must determine which information is to be disclosed to the data subject (on the (dis\nobreakdash-)advantages of this solution, see \citealt{Dubovitskaya2025}).

For consumer credit agreements, there is a specific Right to Explanation in Art.\ 18(8)(a) CCD 2023 (2023/2225/EU), which becomes applicable on November~20,~2026 following transposition by Member States.
It goes somewhat further than the Right to Explanation under the GDPR: It is sufficient that the creditworthiness assessment involves the use of automated processing of personal data, so that the Right to Explanation applies even if automated data processing is only one component of the assessment. Under the GDPR, by contrast, the decision must be based solely on automated processing (cf.\ Art.\ 22(1) GDPR, to which Art.\ 15(1)(h) GDPR refers).
However, Art.~22(1)~GDPR includes not only
decisions taken without any human involvement (Recital 71 of the GDPR), but also those where human participation is merely formal, for example where a person processes a decision but has no ability to deviate from an automatically generated outcome \cite{Buchner2024}.
In addition, in light of the CJEU's SCHUFA judgment \cite{CJEU2023}, fully automated profiling results may also constitute decisions within the meaning of Art.\ 22(1) GDPR, at least where those results decisively shape the subsequent decision \cite{Dubovitskaya2024}. This expands the category of fully automated decisions significantly.

\begin{table*}[h]
\centering
\caption{Inclusion and exclusion criteria for the systematic literature review.}
\label{tab:criteria}
\small
\begin{tabular}{p{2.5cm}p{5.1cm}p{9cm}}
\hline
\textbf{Topic} & \textbf{Inclusion} & \textbf{Exclusion} \\
\hline
Language &
  English or Undefined &
  Non-English Language \\
Year of publication &
  Publication in 2024 or later &
  Publication before 2024 \\
Type of publication &
  Book or article in a journal or conference &
  Patents \\
Use Case specific &
  Discussion of a group of applications &
  Implementation of one specific (X)AI application \\
Explainability &
  Explicit mention of XAI and related approaches and discussion of suitable methods &
  \begin{minipage}[t]{9cm}
    \begin{itemize}[leftmargin=*, topsep=0pt, partopsep=0pt, parsep=0pt, itemsep=2pt]
      \item Broad remarks about XAI (``in general (not) applicable''), only mentions of standard methods such as LIME and SHAP
      \item XAI only as a side note in a general approach towards AI compliance
      \item Positioning of a very specific approach towards XAI (e.g., knowledge graph
            construction) without proper comparison to other approaches regarding their
            legal benefits and drawbacks
    \end{itemize}
  \end{minipage} \\
Law &
  Engagement with the legal bases on the Right to Explanation, especially Art.\ 15 GDPR
  and/or Art.\ 86 AIA and corresponding case law &
  \begin{minipage}[t]{9cm}
    \begin{itemize}[leftmargin=*, topsep=0pt, partopsep=0pt, parsep=0pt, itemsep=2pt]
      \item Outdated legal bases (e.g., now-abandoned AI liability directive of the EU or a version of the AIA without Art.\ 86)
      \item Focus on other legal frameworks without a detailed alignment with the AIA and/or GDPR
    \end{itemize}
  \end{minipage} \\
\hline
\end{tabular}
\end{table*}

Similar to the Right to Explanation in the CCD, the Right to
Explanation in Art.\ 86 AIA covers both fully and partially automated decision-making, since it concerns decisions ``taken on the basis'' of the AI system's output.
However, this right is subsidiary, meaning it only applies if no equivalent right already exists under other Union law provisions (Art.\ 86(3) AIA).
Furthermore, Art.\ 86 AIA is limited to the use of certain high-risk AI systems (those classified as high-risk under Art.\ 6(2) and Annex III of the AIA, with some exceptions), for example in areas like education, employment, healthcare, creditworthiness assessment, life and health insurance, emergency services, law enforcement, and others.
Art.\ 86 AIA was originally scheduled to apply from August~2,~2026, but the Digital Omnibus Act postponed this to December~2,~2027 (standalone systems) or August~2,~2028 (systems embedded in regulated products).
The EU legislator assumes that high-risk systems can significantly impact a person's health, safety, or fundamental rights; the affected person may therefore request from the deployer a clear and meaningful explanation of the AI system's role in the decision-making process and the main elements of the decision.
What exactly constitutes the ``main elements'' is still unclear, but it is likely the CJEU will interpret this provision in disputes similarly to the Dun \& Bradstreet ruling.

\section{Method}
\label{sec:method}

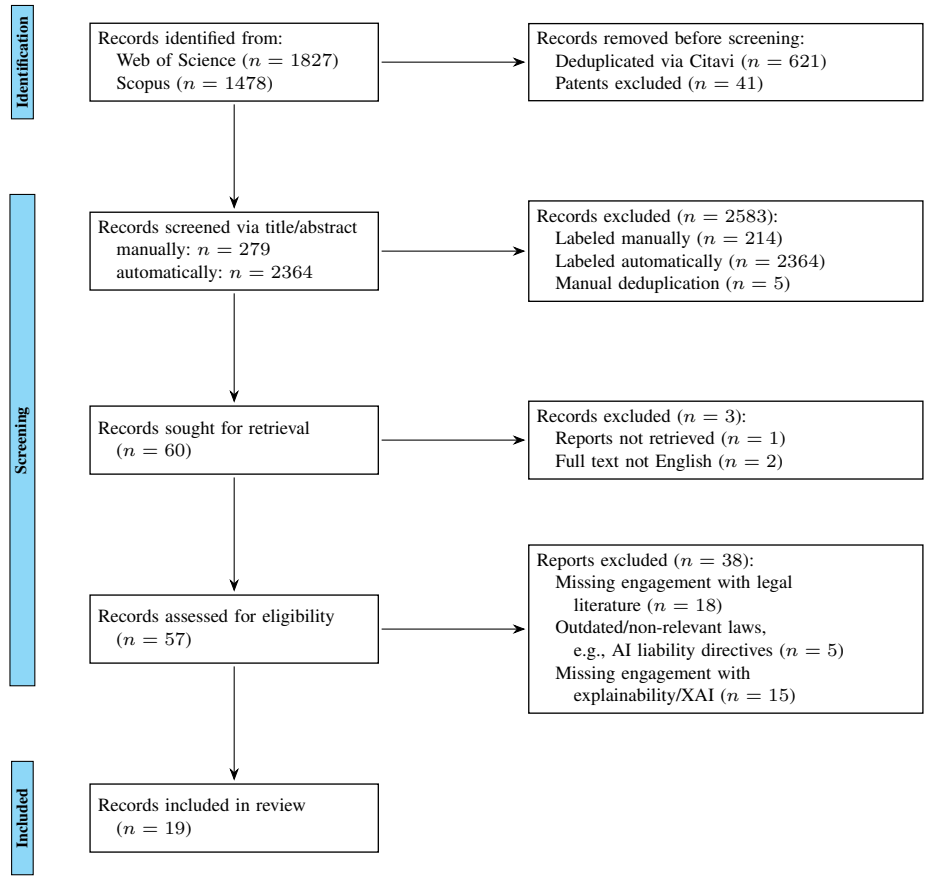
\begin{figure*}[bthp]
\centering
\begin{tikzpicture}[
    mainbox/.style={
        rectangle, draw=black, line width=0.5pt,
        text width=3.6cm, align=left,
        minimum height=0.9cm, inner sep=3pt,
        font=\fontsize{7}{8.5}\selectfont
    },
    exclbox/.style={
        rectangle, draw=black, line width=0.5pt,
        text width=5.0cm, align=left,
        minimum height=0.9cm, inner sep=3pt,
        font=\fontsize{7}{8.5}\selectfont
    },
    phaselabel/.style={
        rectangle, draw=black, fill=cyan!40,
        font=\fontsize{6}{7}\selectfont\bfseries,
        inner sep=2pt, rotate=90, align=center
    },
    arrow/.style={-Stealth, thin, shorten >=1pt, shorten <=1pt}
]

\def\ystep{2.5}  

\node[mainbox] (id) at (0, 0)
    {Records identified from:\\
     \quad Web of Science ($n = 1827$)\\
     \quad Scopus ($n = 1478$)};

\node[mainbox] (screened) at (0, -1*\ystep)
    {Records screened via title/abstract\\
     \quad manually: $n = 279$\\
     \quad automatically: $n = 2364$};

\node[mainbox] (retrieval) at (0, -2*\ystep)
    {Records sought for retrieval\\
     \quad ($n = 60$)};

\node[mainbox] (eligibility) at (0, -3*\ystep)
    {Records assessed for eligibility\\
     \quad ($n = 57$)};

\node[mainbox] (included) at (0, -4*\ystep)
    {Records included in review\\
     \quad ($n = 19$)};

\node[exclbox] (excl_id) at (6.5, 0)
    {Records removed before screening:\\
     \quad Deduplicated via Citavi ($n = 621$)\\
     \quad Patents excluded ($n = 41$)};

\node[exclbox] (excl_screen) at (6.5, -1*\ystep)
    {Records excluded ($n = 2583$):\\
     \quad Labeled manually ($n = 214$)\\
     \quad Labeled automatically ($n = 2364$)\\
     \quad Manual deduplication ($n = 5$)};

\node[exclbox] (excl_retrieval) at (6.5, -2*\ystep)
    {Records excluded ($n = 3$):\\
     \quad Reports not retrieved ($n = 1$)\\
     \quad Full text not English ($n = 2$)};

\node[exclbox] (excl_elig) at (6.5, -3*\ystep)
    {Reports excluded ($n = 38$):\\
     \quad Missing engagement with legal\\
     \qquad literature ($n = 18$)\\
     \quad Outdated/non-relevant laws,\\
     \qquad e.g.,\ AI liability directives ($n = 5$)\\
     \quad Missing engagement with\\
     \qquad explainability/XAI ($n = 15$)};

\node[phaselabel, minimum width=1.5cm]
    at (-2.8, 0) {Identification};

\node[phaselabel, minimum width=6.5cm]
    at (-2.8, -2*\ystep) {Screening};

\node[phaselabel, minimum width=1.5cm]
    at (-2.8, -4*\ystep) {Included};

\draw[arrow] (id)          -- (screened);
\draw[arrow] (screened)    -- (retrieval);
\draw[arrow] (retrieval)   -- (eligibility);
\draw[arrow] (eligibility) -- (included);

\draw[arrow] (id.east)          -- (excl_id.west);
\draw[arrow] (screened.east)    -- (excl_screen.west);
\draw[arrow] (retrieval.east)   -- (excl_retrieval.west);
\draw[arrow] (eligibility.east) -- (excl_elig.west);

\end{tikzpicture}%
\caption{PRISMA flow diagram of the literature search and selection process.}
\label{fig:prisma}
\end{figure*}

We conduct a systematic literature review (SLR) following the PRISMA methodology \cite{Page2021}, surveying literature on the implementation of the Right to Explanation via XAI within EU governance frameworks.
Two bibliographic databases were searched: Web of Science and Scopus, using a search string combining explainability-related terms with EU legal instrument names.
After deduplication and exclusion of patents, 2643 records remained.
Title/abstract screening was conducted using ASReview \cite{vandeSchoot2021}; after an additional validation check with no additional records found, 60 articles proceeded to full text retrieval, of which 57 were assessed for eligibility by two reviewers---one with a technical XAI background, one with a legal background---applying the predefined inclusion and exclusion criteria in Table~\ref{tab:criteria}.
Of these records, 19 matched all inclusion criteria and were included in the final literature review.
The remaining 38 were excluded for insufficient legal grounding (18 with limited coverage of AIA and/or GDPR, 5 based on---for our research question---irrelevant or outdated laws) or insufficient XAI content (15).
A full account of the search strings, screening procedure, stopping criterion, and validation sample is provided in Appendix~\ref{app:method}; a PRISMA flow diagram
is provided in Figure~\ref{fig:prisma}.

\section{Literature Results}
\label{sec:findings}

Of the papers surveyed, only a very limited number (19) were deemed relevant to our
research objective
\cite{Chung2024,Colmenarejo2025,Engelfriet2025,Feretzakis2025,Fresz2024,Gallese,Gorski2025,Grabowicz2023,Hauselmann2025,Juliussen,Kastner,Metikos2024,Metikos2025,Moreira,Palazzo2025,Pavlidis2024,Sapienza,Skorjanc2025,State2025}.
These articles are discussed in the following, both from a legal and technical perspective, insofar as these perspectives can be disentangled for the Right to Explanation.

\subsection{Classification of XAI Methods}
\label{sec:xai-classification}

While the laws discussed provide some guidance on explanation requirements---particularly regarding the main factors addressees and purpose of explanations (see Section~\ref{sec:operationalization})---clear technical requirements remain lacking, starting with the dimensions along which AI models and XAI methods should be viewed.
The surveyed literature reflects three approaches to this classification.

\textbf{I)} Papers discussing the need for explanations in general, often in a more general approach for AI compliance, were
excluded, as this need is evident from the law
(Section~\ref{sec:right-to-explanation}).

\textbf{II)} Most papers separate XAI methods into broad categories, most commonly
post-hoc versus ante-hoc or interpretable methods \cite{Chung2024,Gallese,Grabowicz2023,Juliussen,Metikos2024,Metikos2025,Moreira}.
Post-hoc methods generate explanations after a black-box model has been trained, typically producing local (decision-specific) explanations.
Ante-hoc or interpretable models---such as decision trees or linear models---are intelligible by construction, as long as they remain limited in size.
Within these categories, especially for post-hoc methods, many different explanation methods with different theoretical foundations and properties, but possibly the same explanation format, exist.
Such methods can provide differing explanations of the same model and decision, which is known as the ``disagreement problem'' in XAI \cite{Krishna2022}.
The separation into post-hoc and ante-hoc is in some works connected to the definitions of explainability and interpretability, where both are sometimes used interchangeably or to denote different concepts.
Usually, interpretability denotes interpretable models that can be understood by themselves (also called intelligibility) \cite{Gallese,Gorski2025,Pavlidis2024} as in ``interpretable ML'' or somewhat differently as ``[t]he degree to which a human can understand the cause of a decision'' \cite{Miller2017} as used in \cite{Juliussen}.
In contrast, explainability denotes additional information necessary to understand a model or decision \cite{Gorski2025,Pavlidis2024}, possibly by creating a second model (\citet{Gallese} based on \citet{Rudin2019}) or just to describe local explanations \cite{Juliussen}.
Some papers also use a more specific separation of approaches via their explanation format, e.g., counterfactual explanations or feature importance explanations \cite{Engelfriet2025,Palazzo2025,Skorjanc2025}.
These explanation formats provide more detail to discuss which sort of explanations might satisfy the relevant requirements.
But since many different methods to create explanations of a given format exist, some information might also be lacking---e.g., whether explanations need to perfectly conform to the model in question (a property called Correctness, see below), or whether some kind of approximation might be sufficient.

\textbf{III)} The most specific approaches employ lists of desiderata \cite{Colmenarejo2025,Fresz2024}.
\citet{Fresz2024} draw on the Co-12-properties from \citet{Nauta2022} augmented by five process-properties, while \citet{Colmenarejo2025} derive five desiderata from three technical publications \cite{Chen2022,ChristophMolnar2022,Guidotti2019} and related works.
The different underlying bases impede direct comparison, and several desiderata---such as a property called Correctness by \citet{Fresz2024} or Fidelity/Faithfulness by \citet{Colmenarejo2025}---lack established quantification methods \cite{Moreira,Nauta2022,Tomsett2019}.
For an overview of the properties used in these two publications, see Appendix~\ref{app:properties}.
Both works frame requirements as qualitative soft requirements, either acknowledging partial fulfillment \cite{Colmenarejo2025} or explicit tradeoffs \cite{Fresz2024}.
Such frameworks are closest to operationalization, but risk searching for a level of detail that law does not---and by design should not---provide; as technical standards are meant to provide the technical details for compliance.
Nonetheless, such works could influence future standard development.
\citet{Juliussen} argues that---given evolving XAI capabilities---legal explanation requirements scale with the state of the art, so that AI systems only need to meet currently achievable explanation standards.
In our view, this position is untenable: Where law requires intelligible explanations and a black-box model cannot provide them, the system must not be deployed.
This can be argued for on different grounds: Usually, the influential paper by \citet{Rudin2019} with the rather descriptive title ``Stop Explaining Black Box Machine Learning Models for High Stakes Decisions and Use Interpretable Models Instead'' is cited \cite{Feretzakis2025,Fresz2024,Gallese,Grabowicz2023,Kastner,Metikos2024,Metikos2025}, but also the inability to detect discrimination (\citet{Gallese} based on \citet{Rudin2019}, \citet{Babic2021} and \citet{Vale2022}) or to provide good medical practice (\citet{Gallese} based on \citet{Duran2021}), which cannot be guaranteed using opaque models.

A further common distinction separates Large Language Models (LLMs) from other AI models \cite{Feretzakis2025,Gorski2025}, given that standard XAI methods may perform poorly on large models or unstructured data.
Papers mainly addressing LLMs broadly characterize LIME and SHAP as sufficient for non-LLM models \cite{Feretzakis2025} or at least for provider-deployer relationships \cite{Gorski2025}---a claim directly contradicted by the Addressee/Purpose Framework (Section~\ref{sec:framework}).
The legal texts draw no explicit boundary between LLMs and other models, though the General-Purpose AI provisions (Art.~51 AIA~et~seq.) are primarily aimed at LLMs and image-generation models.

\subsection{Limitations of XAI Methods}
\label{sec:xai-limitations}

Many of the surveyed papers, irrespective of the level of detail they use to describe XAI
methods, agree that explanations generated via post-hoc XAI may seem reliable and
convincing, but are not easily understood by end users or do not necessarily conform to the
model in question or the decision made
\cite{Chung2024,Colmenarejo2025,Fresz2024,Feretzakis2025,Gorski2025,Kastner,Metikos2024,Pavlidis2024,Sapienza,State2025,Skorjanc2025,Engelfriet2025,Palazzo2025}.
\citet{Engelfriet2025} even proposes the term ``principal reason fallacy,'' defined as
``the belief that every algorithmic decision can be traced back to a single, stable, human-interpretable rationale.''
Without this assumption, XAI methods in their current form might be deemed to fail, as decisions that cannot be traced back to a human-interpretable rationale also cannot truthfully be explained via such a rationale---which might be exactly
the type of causal explanation expected by law \cite{Hauselmann2025}.

As a result of the difficulties in understanding XAI---especially in combination with the
overconfident marketing of XAI capabilities---\citet{Chung2024} argue
that the legislative pressure to provide explanations will result in companies using and
trusting suboptimal XAI explanations. To circumvent this problem, they suppose that detailed
regulations could define and evaluate standards for XAI, otherwise ``ad hoc implementations
could lead a right to explanation astray'' \cite{Chung2024}. A similar notion is provided by
\citet{Moreira}, who argue that ``regulatory requirements imply
verification of the XAI methods, guaranteeing their quality.
Otherwise, ignoring the quality of explainability is equivalent to ignoring this transparency requirement and may even be harmful, as a false explanation may cause more damage than no explanation at all.''

\subsection{Legal Grounding in the Surveyed Literature}
\label{sec:legal-grounding}

Interestingly, the new CCD, with its Right to Explanation, is only mentioned twice in the surveyed papers~\cite{Engelfriet2025,Skorjanc2025}.
This applies even to contributions that otherwise engage with consumer protection \cite{Colmenarejo2025}, cite credit denial as an example of an AI-supported decision \cite{Fresz2024,Gorski2025,Juliussen,Metikos2025}, or are based on a study explicitly concerned with explanations for credit decisions, and thus centrally address the topic of credit decision-making \cite{State2025}.
Outside the reviewed corpus, there are  contributions that do discuss the CCD; however, they do not sufficiently engage with the AIA or with XAI and had to be excluded on that basis.
Conversely, the papers that focus on GDPR and AIA commonly do not refer to the CCD, even though a Right to Explanation is a common element across these instruments.

Regarding the GDPR, the literature presents a heterogeneous and overall rather blurred
picture. Some papers do not address the GDPR at all \cite{Kastner}, while others mention it
only in very general terms, without reference to specific provisions and without engaging
with a Right to Explanation, because they treat explainability as a regulatory principle
without dogmatically linking it to particular provisions of the GDPR \cite{Pavlidis2024}.

Most papers address the GDPR-based Right to Explanation but 
ground it imprecisely.
This right is often derived from Art.\ 22 GDPR, sometimes in conjunction with Recital 71, instead of being grounded in Art.\ 15(1)(h)~GDPR
\cite{Colmenarejo2025,Chung2024,Gallese,Feretzakis2025,State2025}.
In some cases, reference is made exclusively to Recital 71 \cite{Grabowicz2023}, despite recitals not being binding provisions.
Other works describe the legal basis of the Right to Explanation in the GDPR as unclear \cite{Gorski2025}, while simultaneously characterizing the right itself as controversial \cite{Chung2024,Gorski2025,Moreira}.
Still other papers cite both Art.\ 13--15 GDPR and Art.\ 22 GDPR in conjunction with Recital 71 as possible legal bases, without clearly distinguishing between them \cite{Metikos2025,Metikos2024}.
Only a small number of works provide a correct norm-dogmatic grounding of the Right to Explanation \cite{Fresz2024,Juliussen,Sapienza,Hauselmann2025,Skorjanc2025}.

The ambiguities described above may create the impression that the GDPR does not provide a Right to Explanation, or only a weak/limited version, although this is not the case.
The Right to Explanation has a firm legal basis in the GDPR, namely in Art.\ 15(1)(h), a point that has been settled at the latest since the CJEU's Dun \& Bradstreet judgment \cite{CJEU2025}.
An accurate account of the legal framework is particularly important at this juncture to avoid the mistaken assumption that the Right to Explanation under the GDPR exists only in conjunction with the corresponding right under the AIA, or that it is weaker than the latter (see, e.g., \citealt{Gorski2025}).

As discussed in Section~\ref{sec:right-to-explanation}, the CJEU's Dun \& Bradstreet judgment is of central importance to the interpretation of the Right to Explanation under the GDPR.
Despite the Advocate General's opinion in this case foreshadowing this development in 2024~\cite{AGopinion2024}, its uptake in recent literature is rather slow.
Even among the papers published in 2025 and 2026 (14 papers in total), only five take this judgment into account, although it was delivered early in the year, on February 27.
Notably, the newer the papers, the more likely they were to include this information, pointing towards the slowness of scientific publishing being the main reason the judgment was not incorporated earlier.

Art.\ 86 AIA is not mentioned in some papers because they focus on the GDPR
\cite{Feretzakis2025,State2025}.
Other contributions may not refer to this provision because it was only added to the text of the Regulation in 2024, shortly before its adoption.

In the majority of the contributions (13 papers), Art.\ 86 AIA is mentioned
\cite{Colmenarejo2025,Gorski2025,Grabowicz2023,Juliussen,Kastner,Metikos2024,Metikos2025,Sapienza,Moreira,Engelfriet2025,Hauselmann2025,Palazzo2025,Skorjanc2025}.
However, some authors do not engage with the Right to Explanation because they conceptualize explanations from a different analytical perspective, for instance as an instrument of institutional accountability \cite{Grabowicz2023}, as a functional device for liability attribution \cite{Kastner}, or as a procedural precondition for effective judicial protection \cite{Metikos2024}.

By contrast, the papers that examine the Right to Explanation in the context of the AIA correctly derive from Art.\ 86(1) AIA a Right to Explanation of individual decisions \cite{Colmenarejo2025,Gorski2025,Juliussen,Metikos2025}.
While these authors occasionally acknowledge that the wording, in particular the reference to the ``role of the AI system in the decision-making process,'' could theoretically support alternative interpretations, they nonetheless endorse a decision-specific reading \cite{Metikos2025}.

Only a limited number of contributions explicitly analyze the relationship between
Art.\ 15(1)(h) GDPR and Art.\ 86 AIA, notably newer ones
\cite{Juliussen,Hauselmann2025,Skorjanc2025}. Where the two provisions overlap, Art.\ 86
AIA is subsidiary to the corresponding right under the GDPR, pursuant to Art.\ 86(3) AIA.
At the same time, it is emphasized that Art.\ 15(1)(h) GDPR does not cover all conceivable scenarios.
As a result, Art.\ 86 AIA retains an autonomous and non-redundant field of
application.

\subsection{Requirements on Explanation Form}
\label{sec:design-requirements}

It is a positive aspect that the requirements for the design of explanations are regularly
discussed in the papers under review
\cite{Colmenarejo2025,Fresz2024,Engelfriet2025,Palazzo2025,Skorjanc2025,Hauselmann2025}
and, in part, explicitly linked to legal standards. In interpreting Art.\ 86 AIA, Recital 171 is frequently invoked, as it calls for clear and meaningful explanations on the basis of
which the affected person can exercise their rights. From this, it is correctly inferred that
an explanation must contain ``more'' than merely ``values of features and their importance
scores'' and must therefore be understandable to the affected person without the involvement
of an AI expert \cite{Gorski2025,Metikos2025,Hauselmann2025}. \citet{Hauselmann2025} states that ``If information \ldots{} does not relate to a particular
decision, it cannot be meaningful for data subjects in light of the remedies the GDPR
provides'', tying ``meaningful'' to local explanations. Furthermore, they argue that humans
usually search for causality in explanations, and explanations should ``explain which factors
affected the final ADM [Automated Decision-Making]'' to ``determine which factors they must
challenge to change the ADM by obtaining human intervention, expressing their point of view,
and contesting the decision as foreseen in Article 22 (3) GDPR.'' Such explanations are
called causal here, although they do not fit the strict interpretation of causality but more
the notion of ``actionability'' of \citet{Wachter2018}.

Additionally, \citet{Hauselmann2025} notes that ``regarding the
concept `meaningful information' the CJEU acknowledges that the different language versions have various, different meanings'', which are thought to be complementary, but complicate the implementation of such requirements.

Overall, in the context of the GDPR, however, a strong linkage to the existing legal requirements is still missing.
For instance, in a study that sought to examine the requirements for explanations under the GDPR, test subjects were presented with SHAP visualizations as explanations for certain credit decisions \cite{State2025}.
The test subjects, who were lawyers and thus technical laypeople, had difficulties understanding these explanations and unanimously expressed a preference for textual rather than graphical explanations.
The authors of the study conclude that commonly used XAI explanations are often ill-suited for the exercise of data subject rights because they are perceived as difficult to understand and substantively incomplete.
What is missing, however, is the observation that such explanations fail to meet the requirements of Art.\ 12(1) GDPR, although the authors refer to this provision at several points.
It requires that information, including information provided pursuant to Art.\ 15 GDPR, be communicated in a concise, transparent, intelligible, and easily accessible form, using clear and plain language.
The CJEU also relied on this provision in its Dun \& Bradstreet judgment \cite{CJEU2025}.
The notion of ``language'' in Art.\ 12 GDPR must be understood as referring to natural language used in everyday, non-technical contexts.

The papers under review further address the tension between accuracy and intelligibility, arguing that an explanation must strike a balance between ease of understanding and a sufficient level of detail \cite{Colmenarejo2025}.
Here again, a linkage to legal doctrine is instructive, as Art.\ 12(1) GDPR is likewise understood to give rise to two potentially conflicting requirements.
On the one hand, the requirement of accuracy obliges the controller to provide substantively precise information; on the other hand, the requirement of intelligibility demands that the information be presented in such a way that the data subject can actually comprehend it without excessive cognitive or temporal effort \cite{Backer2024}.
It has therefore been proposed to resolve this tension by granting the controller a margin of discretion in the choice of the form of presentation, with a violation of Art.\ 12(1) GDPR being assumed only in cases of manifestly inaccurate or unintelligible information \cite{Backer2024}.

\subsection{Purpose and Content of Explanations}
\label{sec:explanation-objectives}

Many authors agree that the law specifies the objectives that explanations are meant to
achieve; these objectives, in turn, determine the content of explanations and, potentially,
the appropriate explanatory methods
\cite{Colmenarejo2025,Fresz2024,Juliussen,Metikos2025,Sapienza,Engelfriet2025}. Under
Art.\ 15(1)(h) GDPR, for example, explanations are intended to enable the data subject to
exercise the right to contest the decision pursuant to Art.\ 22(3) GDPR. Accordingly, such
explanations must provide the data subject with the information necessary to raise objections
to the correctness of the decision.

Unlike the GDPR, the AIA does not provide for a right of the affected person to contest a decision.
However, Recital 171 states that explanations under Art.\ 86 AIA are intended to provide the affected person with a basis for exercising ``their rights.''
In the absence of a harmonized Union rule, the nature of these rights is determined by the applicable national law.
These may include, for example, the right to challenge administrative decisions in public law or the right to claim damages under the private law of a Member State. In order to exercise such rights, the affected person must, similarly to the GDPR context, be placed in a position to challenge the decision.

For this purpose, most technical explanations (mainly referring to the output of XAI methods) are not sufficient, as they are merely descriptive.
Legal explanations might also need to provide information on how to contest decisions, the general functioning of AI systems, the design decisions influencing the final system, the roles of humans involved in the overall processing of data and normative information on relevant regulation and applicable norms \cite{Engelfriet2025,Skorjanc2025,Almada2025,Kolarova2026,Moreira,Hauselmann2025}.
Notably, the publications describing such additional information needs also tend to include multiple technical explanations in their suggestions on how to present explanations, e.g.,~\citet{Engelfriet2025} (see below).

The objective an explanation is required to fulfill must therefore be determined through the interpretation of the specific legal provision at issue, and these objectives may vary considerably.
In product safety and product liability law for instance, the focus is not on contesting machine learning decisions but on uncovering product defects \cite{Fresz2024}.
In this context, reference should also be made to the work of \citet{Kastner}, which addresses liability for AI-mediated harm. The authors argue that
liability law requires a distinction between three questions: (1) which inputs caused the harmful output and who is responsible for it; (2) which functional components, units, or circuits within the model were causally decisive; and (3) which training data or design decisions led to the harmful behavior.
According to the authors, each of these questions calls for a distinct explanatory approach, namely classical XAI methods for questions of type (1), mechanistic interpretability
(interpreting the functioning of an AI model based on specific ``circuits'') for questions of
type (2), and analysis of the system's overall history for questions of type (3)
\cite{Kastner}. It remains notable, however, that data-centric XAI strategies like
input-influence methods are not considered for type (3) questions in this framework.
A similar framework is described by \citet{Engelfriet2025}, who argues that explanations should entail four parts: (1) statistical transparency, e.g., via XAI; (2) distributive contextualization through comparison to similar data points; (3) normative linkage, somewhat similar to the concept of ``justification'' (see Section~\ref{sec:new-concepts}), to argue ``why an outcome is acceptable under applicable legal or ethical norms''; and (4) contrastive actionability, e.g., via counterfactual explanations.

\subsection{New Concepts}
\label{sec:new-concepts}

Some of the papers we reviewed introduce, in addition to the concept of an ``explanation,''
the notion of ``justification'' \cite{Colmenarejo2025,Gorski2025,Engelfriet2025}. However,
the papers employ the concept of justification in slightly different ways.
\citet{Colmenarejo2025} use it to refer to the lawfulness of
the decision; justification, in their account, shows why a decision is acceptable in light
of applicable norms, principles, and the facts of the case, as opposed to an explanation,
which merely sets out how the system arrived at the outcome in input--output terms.
\citet{Gorski2025}, by contrast, understand justification more as a
form of reasoning that seeks to establish the lawfulness of a decision, without necessarily
implying that the decision is objectively lawful. \citet{Engelfriet2025} stresses that
technical explanations only provide statistical correlations, but legal explanations also
have normative value and thus provide justifications of why a decision is (or might be)
legitimate (similar to \citealt{Kolarova2026}). These authors frame justification as a
counterweight to purely technical explanations that are unintelligible to laypeople;
ultimately, their concern is therefore with the form of explanations (see above).

From a legal perspective, it is nevertheless crucial to recognize that Art.\ 15(1)(h) GDPR,
Art.\ 18(8)(a) CCD, and Art.\ 86 AIA concern explanations rather than
the lawfulness of decisions as such. The purpose of an explanation is solely to make the
essential reasons underlying a decision comprehensible to the affected person. An explanation
is not, by itself, proof that the decision is lawful. Rather, it is intended to provide the
information necessary to enable a review of the decision's lawfulness.
This raises the question whether the concept of ``justification'' adds analytical value in the context of explanations or instead risks generating confusion.

A different notion of legal explanations is provided by \citet{Sapienza}.
They decompose explanations based on the addressed explainee:
``deployer-oriented (ensuring system transparency for appropriate use),
compliance-oriented (documentation for regulatory adherence), individual-empowering (rights
to contest AI-supported decisions), and oversight-oriented (tools enabling meaningful human
control)'' \cite{Sapienza}. With these, they provide two definitions: Knowability as an
overall framework encompassing different forms of explainability (comparable to transparency
in other literature) and Explainability as the disclosure of elements regarding the inner
workings of AI algorithms.
Starting from the common distinction of local and global explanations, they propose a new categorization of explanations into enabling, certifying/ confirming and actionable explanations, aiming at giving the user a basis to decide whether to use an AI system for a decision, verifying or aligning an AI system with regulatory or judicial requirements, or empowering an individual to object to a
decision \cite{Sapienza}.
While such an approach is interesting from a legal perspective,
especially since the different kinds of explanations are argued to be necessary in
combination, it adds complexity (see above) and lacks detail regarding its
applicability in practice, as most XAI methods do not specify their purpose or validate their explanations with regard to the three explanation types defined by \citet{Sapienza}.

A similar notion is provided by \citet{Fresz2024}, as they separate
the legal requirements into decision-centric---from the perspective of decision makers performing a plausibility check or decision recipients using the Right to Explanation---and model-centric---for product safety or product/tortious liability \cite{Fresz2024}.
These are mapped to XAI properties, resulting in different properties the authors deem necessary for each use case (see Section~\ref{sec:xai-classification}).


Starting from the separation of explanation and justification, \citet{Colmenarejo2025} define legal desiderata on explanations: normativity (adapting explanations and justifications to the specific field of law); purposefulness (does the explanation or justification suit the purpose it shall satisfy); truthfulness (providing accurate, truthful and complete information); intelligibility; accessibility (of explanations or justifications) \cite{Colmenarejo2025}.
They note that these legal desiderata are qualitative and do not allow for a quantitative analysis.
As such, these principles would need a use-case specific manual assessment---and more specific definitions to clarify the concepts themselves---to determine whether an explanation sufficiently satisfies them.
This approach, while useful as a heuristic, derives its desiderata primarily from technical literature rather than from specific legal norms, and does not establish the independence of form and content requirements based on the structure of the relevant provisions.



\section{Operationalization of Explanation Requirements}
\label{sec:operationalization}

The previous literature analysis reveals two related
problems: a structural conflation of form and content requirements, and an
operationalization gap whereby even correctly identified requirements cannot
be translated into verifiable XAI specifications.
Section~\ref{sec:framework} provides the Addressee/Purpose Framework as the
analytical foundation; Section~\ref{sec:blueprint} introduces a four-phase
blueprint that structures the open research agenda in Section~\ref{sec:agenda}.

\subsection{The Addressee/Purpose Framework}
\label{sec:framework}

The literature analysis in Section~\ref{sec:findings} reveals a conflation
of two legally distinct dimensions that govern explanation requirements under EU law.
We conceptualize these as the Addressee/Purpose Framework, which provides
practitioners with a structured design principle for legally compliant XAI deployments.
While \citet{Sapienza} categorize by addressee type and \citet{Fresz2024} distinguish by legal purpose, neither treats these as independently necessary legal dimensions anchored in distinct provisions---a structure our framework makes explicit.

The \textbf{addressee} of an explanation determines its required \textbf{form};
the \textbf{purpose} of an explanation---derived from the specific legal provision
triggering the obligation---determines its required \textbf{content}.
This distinction maps directly onto the structure of the relevant legal instruments.
Art.~12(1) GDPR governs form: Explanations must be provided ``in a concise, transparent, intelligible and easily accessible form, using clear and plain language.''
By contrast, the substantive provisions---Art.~15(1)(h) GDPR, Art.~86 AIA, and Art.~18(8)(a) CCD---govern content: They specify what information must be provided so that the addressee can exercise their rights.
The CJEU confirmed in its Dun \& Bradstreet judgment that both dimensions must be fulfilled simultaneously and independently, a point not sufficiently acknowledged in previous literature.
Examples of possible addressees and
explanation purposes are set out in Table~\ref{tab:addressees-purposes}.

\begin{table}[t]
\centering
\caption{Example categories for explanation addressees and corresponding purposes.}
\label{tab:addressees-purposes}
\small
\begin{tabular}{p{3.8cm}p{3.8cm}}
\hline
\textbf{Explanation Addressee} & \textbf{Explanation Purpose} \\
\hline
\begin{minipage}[t]{3.8cm}
  \begin{itemize}[leftmargin=*, topsep=2pt, partopsep=0pt, parsep=0pt, itemsep=2pt]
    \item Non-expert user
    \item Data subject (GDPR)
    \item Affected person (Art.\ 86 AIA)
    \item A child (see Art.\ 12(1) GDPR)
  \end{itemize}
\end{minipage}
&
Generally: contestation of the automated decision (Art.\ 22(3) GDPR, Art.\ 86 and Recital
171 AIA) \\
\hline
Semi-expert user, e.g., personnel of a regulatory authority &
  Review of the AI system's compliance with regulatory requirements \\
\hline
Expert user (e.g., qualified personnel of the deployer) &
  E.g., to enable human oversight or compliance with applicable legal requirements
  during the development of an AI system \\
\hline
\multicolumn{1}{c}{\large\(\Downarrow\)} & \multicolumn{1}{c}{\large\(\Downarrow\)} \\
\multicolumn{1}{c}{\textbf{Form of the explanation}} &
  \multicolumn{1}{c}{\textbf{Content of the explanation}} \\
\hline
\end{tabular}
\end{table}

\subsubsection{Practical Consequences}
The distinction carries non-trivial practical consequences.
A technically correct explanation may satisfy the content requirement while failing the form requirement for a non-expert data subject, since many explanations presuppose statistical literacy that data subjects generally lack (cf.~\citealt{State2025}).
Conversely, a plain-language narrative may satisfy the form requirement while omitting the contrastive or counterfactual information needed to contest the decision, thereby failing the content requirement.
Neither dimension can substitute for the other, and no technical solution that addresses only one of the two can be considered legally compliant.

We provide an illustrative example in the following to highlight the joint importance and interplay of these two dimensions.
A credit decision under Art. 15(1)(h) GDPR requires (a) form: plain-language text (potentially aided by statistical visualizations), per Art. 12(1) GDPR as interpreted in Dun \& Bradstreet; and (b) content: contrastive or counterfactual information sufficient to enable contestation under Art. 22(3) GDPR.
A pure SHAP visualization satisfies neither dimension: It fails form (not plain language for a non-expert) and likely fails content (descriptive rather than actionable).
This directly contradicts claims in the surveyed literature that LIME and SHAP are sufficient for non-LLM contexts---a contradiction the Addressee/Purpose Framework makes structurally visible.
An automatically generated text summary satisfies form but may still fail content if it omits contrastive information.
As of now, it remains unclear whether both dimensions can be satisfied by a single explanation, as a truly correct explanation might just be too complex to be intelligible or might not map onto human-understandable reasons~\cite{Engelfriet2025}.

Despite a perfect solution for explanation form and content not existing, some interim recommendations can be given to build current XAI systems:
\begin{enumerate}
    \item \textbf{Modular pipelines.} XAI pipelines should be modular, with multiple addressee-specific explanations from a shared model, fitting to the related purpose.
    \item \textbf{Empirical validation of form.} Explanation form should be empirically validated for intelligibility through user studies---not merely verified for technical properties.
    \item \textbf{Purpose-based method selection.} XAI methods should be chosen based on the legal purpose of the explanation.
    \item \textbf{Documentation of design choices.} Design choices in explanations should explicitly be documented to show which addressee category and legal purpose each explanation component is designed to serve (e.g., counterfactual explanations for the ``contest'' purpose).
    \item \textbf{Relying on up-to-date information.} Due to delays in scientific publishing, practitioners should rely not only on scientific papers, but also on more up-to-date information.
\end{enumerate}

\subsection{A Four-Phase Blueprint for Explanation Operationalization}
\label{sec:blueprint}

The Addressee/Purpose Framework identifies \emph{what} legally compliant
explanations must satisfy. The four-phase blueprint describes \emph{how} to
translate this into practice, providing a structured process from determining
applicable requirements to documenting whether and how they are met.

\paragraph{Phase~1 --- Identification of Applicable Requirements.}
The first phase establishes which legal instruments impose explanation
obligations in a given deployment context. This requires resolving scope
questions (e.g., does Art.~22(1) GDPR apply? Is the system high-risk under
Annex~III AIA?), determining precedence where instruments overlap
(Art.~86(3) AIA makes the AIA right subsidiary to equivalent GDPR rights),
and identifying the specific legal purpose triggered by each provision.

\paragraph{Phase~2 --- Breaking Down Requirements into Quantifiable
Sub-Requirements.}
Phase~2 translates the qualitative legal standards identified in Phase~1 into
measurable sub-requirements with verifiable thresholds.
Standards such as ``meaningful'' (Dun \& Bradstreet) or
``intelligible'' (Art.~12(1) GDPR) are technology-agnostic by design;
operationalization requires defining empirically testable criteria for
each---for instance, minimum comprehension rates for intelligibility or
required information types for contestation.
In line with the Addressee/Purpose Framework, form and content
sub-requirements must be derived and measured.

\paragraph{Phase~3 --- Evaluation of XAI Methods Against Sub-Requirements.}
Once sub-requirements are established, available XAI methods can be
systematically evaluated against them. This evaluation must be legally
aware: Assessing abstract technical properties in isolation is insufficient;
outputs must be tested against the sub-requirements specific to the legal
context and addressee in question. Form and content compliance must be
assessed, and the disagreement problem---where different methods
yield conflicting explanations of the same decision---must be managed before
any output can be relied upon as legally adequate.

\paragraph{Phase~4 --- Argumentation of Tradeoffs and (Non\nobreakdash-)Fulfillment.}
The final phase requires practitioners to document and legally defend the
extent to which explanation requirements are met, including cases of partial and non-fulfillment.
Where tradeoffs are unavoidable, design choices must be argued transparently and in a manner amenable to regulatory review.
Where no compliant solution exists, Phase~4 requires explicit acknowledgment that the system must not be deployed in the relevant context.
Standardization bodies have a critical role here by providing harmonized,
verifiable conformance criteria.

\section{Open Research Agenda}
\label{sec:agenda}

The four-phase blueprint makes explicit where operationalization is currently
blocked.
The six research questions below correspond to the phases at which progress
is most urgently needed; they constitute an agenda addressed
to the XAI and legal-informatics communities jointly.

\subsection{Phase~1 --- Identifying Requirements}
\label{sec:agenda-phase1}

\paragraph{RQ\,1 --- Interaction between Art.~15(1)(h) GDPR and Art.~86 AIA
in concurrent deployments.}
As established in Section~\ref{sec:legal-grounding} and confirmed by only a
small subset of the surveyed literature \cite{Juliussen,Hauselmann2025,Skorjanc2025},
the interplay between these provisions remains underspecified.
Systematic analysis is needed to determine which legal explanation obligations apply in which deployment settings---particularly for high-risk AI systems that also process personal data, where both instruments may be simultaneously
applicable---and how conflicts or gaps between the two regimes should be
resolved in practice.

\subsection{Phase~2 --- Quantifying Sub-Requirements}
\label{sec:agenda-phase2}

\paragraph{RQ\,2 --- Operationalizing the ``meaningful'' standard.}
The CJEU in Dun \& Bradstreet requires explanations to be ``meaningful, i.e., useful, relevant, important and easily understandable.''
No empirically validated criteria currently exist for determining whether an XAI-generated explanation meets this standard for a non-expert data subject.
Future work should develop and validate operationalizable definitions for \emph{meaningfulness} that are both technically measurable and legally defensible---for instance, through user studies using legally realistic decision scenarios (credit, employment, healthcare) with participants from the relevant demographic groups.

\subsection{Phase~3 --- Evaluating XAI Methods Against Sub-Requirements}
\label{sec:agenda-phase3}

\paragraph{RQ\,3 --- Legally aware XAI evaluation benchmarks.}
Current evaluation metrics focus on technical properties such as Correctness
or Completeness (see Appendix~\ref{app:properties}), none of which maps directly to legal adequacy.
Benchmark datasets and evaluation protocols are needed that assess whether XAI
outputs enable a non-expert to identify concrete grounds for contesting a
decision, anchored in specific legal provisions rather than abstract desiderata.
Such benchmarks should include realistic decision scenarios and report
results across demographically diverse participant samples to surface potential
disparate impacts of explanation quality.

\paragraph{RQ\,4 --- Scope and Correctness of explanations.}
As noted in Section~\ref{sec:explanation-objectives}, the scope of what needs
to be explained is rather unclear, especially since technical ``explanations''
mainly focus on XAI method output, whereas legal explanations might entail
information such as human involvement in decision-making and normative grounding.
Additionally, different post-hoc XAI methods applied to the same model and
decision can yield conflicting explanations---the ``disagreement
problem''~\cite{Krishna2022}.
In legal contexts this is not merely a technical inconvenience: Contradictory
explanations may violate the requirements of accuracy and truthfulness that
flow from Art.~12(1) GDPR.
Since we cannot expect laypeople to be able to choose between different
disagreeing explanations, this poses the question of how this problem can be circumvented in practice, and how the related systems need to be documented to allow for proper assessment.
Potentially, the evaluation of the Correctness of specific explanation methods could render some methods as insufficiently correct for specific scenarios and others as acceptable (e.g., when based on guarantees~\cite{monke2025}), which would provide practitioners with enough information to choose a legally compliant approach.

\paragraph{RQ\,5 --- Explainability of foundation models in regulated contexts.}
The growing deployment of LLMs in high-risk domains
covered by Art.~86 AIA (education, employment, healthcare) raises unresolved
questions.
Current LLM explanation techniques---attention visualization, chain-of-thought
prompting, integrating knowledge bases---have not been validated against the
``meaningful'' standard for non-expert users, and their Correctness is unclear.
Research should: (a) assess the adequacy of these techniques for legally
mandated explanations; and (b) determine whether the obligation to be able to
provide explanations for AI systems in high-risk use cases entails a de facto
constraint on the use of LLMs (and black-box models in general) in certain
Art.~86 AIA contexts.

\subsection{Phase~4 --- Tradeoffs and (Non-)Fulfillment}
\label{sec:agenda-phase4}

\paragraph{RQ\,6 --- Standardization pathways for Art.~86 AIA.}
No harmonized standard addresses the technical requirements of Art.~86 AIA
explanations, as standardization mandate M/613~\cite{CommissionoftheEuropeanUnion2025} does not cover this provision.
Research should: (a) identify the minimum technical content a compliant
explanation must include for each principal use-case category under
Art.~86 AIA; (b) propose verifiable conformance criteria that can inform
future standardization work; and (c) engage standardization bodies
(CEN-CENELEC JTC~21, ISO/IEC JTC~1/SC~42) with concrete, evidence-based
proposals grounded in the legal analysis presented in this paper.
Without such targeted research, the Right to Explanation under the AIA risks
remaining a regulatory obligation without a technically realizable path to
compliance.

\section{Discussion}
\label{sec:discussion}
This review confirms that EU law imposes explanation obligations, yet no shared reference frame currently translates these into technically operationalizable XAI specifications.
Conceptual ambiguities and misaligned expectations about what law can provide continue to hinder practical synthesis.

A central tension concerns the level of detail law is expected to provide.
EU legislation is deliberately technology-agnostic: It fixes objectives (intelligible, plain-language explanations enabling contestation) rather than methods.
Bridging this gap is the task of technical experts and standardization bodies.
This task is not always reflected in the literature, where at times legal requirements are noted as abstract desiderata and not operationalized further
(see Sections~\ref{sec:xai-classification} and~\ref{sec:new-concepts}), or outputs are only intelligible to experts rather than to the data subjects the law protects (Section~\ref{sec:design-requirements}).

Standards could fill part of this gap.
ISO/IEC~TS 6254:2025, ISO/IEC~12792:2025, and ISO/IEC~DIS~42105 collectively address explainability objectives, transparency taxonomy, and human oversight.
The two published standards (ISO/IEC TS 6254:2025, ISO/IEC 12792:2025) provide shared terminological foundations but lack the operational specificity required for Art. 86 AIA compliance; ISO/IEC DIS 42105 remains unpublished and therefore cannot be assessed for its practical utility.
Notably, standardization mandate M/593 and its replacement M/613 do not cover Art.~86~AIA; no harmonized standard for that provision is forthcoming.
Even existing standards have historically lacked the specificity needed for direct implementation, leaving open where normative technical detail should originate if not from scientific consensus.

The field's immaturity thus risks hardening into a structural standstill: Practitioners and companies await concrete normative guidance before committing to legally compliant XAI architectures, while legal scholars and standardization bodies await technically mature solutions---and further clarifying case law---before specifying enforceable requirements.
Neither side can act decisively without the other, yet the GDPR Right to Explanation is already in force and directly enforceable, with the CCD and AIA rights following in the next years (see Section~\ref{sec:right-to-explanation}). 

Without clearer guidance, practitioners risk falling into one of two extremes: Accepting readily available but technically subpar XAI outputs---embedding a false sense of compliance \cite{Chung2024,Moreira}---or avoiding black-box AI in regulated contexts altogether.
The latter is, on current law and technology, arguably the more easily defensible position with a lower legal risk: Where no method can produce truthful and intelligible explanations as required by law, the system must not be used, directly contradicting the AIA's aim of fostering innovation.

\section{Limitations}
\label{sec:limitations}
The review's scope is intentionally restricted to the EU regulatory landscape (GDPR, AIA, CCD); conclusions may not transfer to other jurisdictions.
The corpus is small---19 papers with substantive dual-domain engagement.
This partly reflects the focused post-2024 temporal scope of the review (see Appendix~\ref{app:method}), but also indicates that relatively few publications substantively integrate both legal and technical perspectives under the current regulatory framework.

Methodologically, results are bounded by specific search terms, an English-language and post-2024 filter (motivated by Art.~86~AIA's late addition to the final AIA text, see Appendix~\ref{app:method}), and coverage of Web of Science and Scopus only.
One paper known to the authors, but not indexed in either database at the time of writing \cite{FrancescoSovrano2025}, was excluded, as it did not yield a substantial additional contribution.
Title/abstract screening by a single reviewer with a technical XAI background introduces potential selection bias; this was mitigated through documented inclusion/exclusion criteria (Table~\ref{tab:criteria}) and collaborative full text review by researchers from both legal and technical backgrounds.
Comparable exclusion rates across both domains suggest effective mitigation, but residual bias cannot be fully ruled out.

\section{Conclusion}
\label{sec:conclusion}

This paper has grounded the Right to Explanation in Art.~15(1)(h)~GDPR, Art.~86~AIA, and Art.~18(8)(a)~CCD, and systematically reviewed the literature bridging these norms with XAI practice.
Of 2643 records from a broad initial search, and 57 full texts surveyed, only 19 papers demonstrated substantive engagement with both domains, showing limited coverage of the current regulation.

The review documents three problematic patterns: Many papers misidentify the GDPR legal basis (Art.~22 rather than Art.~15(1)(h)), do not integrate the CJEU's landmark Dun \& Bradstreet judgment---likely due to publication delays---and conflate the addressee/purpose distinction.
We conceptualize this distinction in the Addressee/Purpose~Framework: The \emph{addressee} of an explanation determines its required \emph{form}; the \emph{legal purpose} of the triggering provision determines its required \emph{content}.
These dimensions are jointly necessary---no existing XAI approach reliably satisfies both, and growing skepticism in recent work suggests this may not be possible~\cite{Engelfriet2025}.
We further introduce a four-phase blueprint for operationalization, showing where each of the six open research questions currently blocks the process from identification of applicable requirements through to argumentation of tradeoffs and (non-)fulfillment.

Progress requires joint advances across all four phases: clarification of applicable legal requirements (RQ~1), operationalization of norms such as ``meaningfulness'' (RQ~2), empirical validation of XAI outputs against legally realistic scenarios with non-expert participants (RQ~3–5), and technically grounded standardization covering Art.~86 AIA (RQ~6).
Without such progress, the Right to Explanation risks remaining a formal obligation without a technically realizable path to compliance---and either companies implementing AI or affected individuals will bear the cost.

\section*{Acknowledgements}
Parts of this paper were created with the help of a company-specific implementation of Claude Sonnet 4.6 R. It was used to create LaTeX code for tables and to refine the drafts of some sections. We thank all reviewers of this paper, who helped to improve it with their valuable feedback.

This paper is supported by the Ministry of Economic Affairs, Skilled Trades and Tourism of Baden-Württemberg within the projects  ``KIRR Real (Reallabor für rechtskonforme KI und Robotik)'', ``Reallabor am KI-Fortschrittszentrum'' and the Testing and Experimentation Facility (TEF) ``AI-Matters'', which is co-funded by the European Union under grant agreement number 101100707.

\section*{Ethical Statements}
\label{sec:ethics}

\subsection*{Ethical Considerations Statement}
\label{ssec:ethical-considerations}

This research is motivated by the protection of individuals who receive consequential automated decisions---e.g., in credit, employment, and healthcare---without adequate explanation.
Some tensions in the work itself require disclosure.

\paragraph{On false compliance.}
Our finding that no current XAI approach reliably satisfies both the form and content dimensions of the Right to Explanation is not an argument for inaction.
The paper explicitly rejects the position that legal explanation requirements scale unconditionally with technical feasibility.
Where intelligible, truthful explanations cannot be produced, it needs to be discussed whether such systems should be deployed at all (Sections~\ref{sec:xai-classification} and~\ref{sec:agenda-phase4}).
We are aware that this finding could be selectively cited to argue the opposite: that compliance is technically impossible and therefore unenforceable.
This is not our intention, as we hold the Right to Explanation to be important and want to contribute to its intention of providing intelligible and correct explanations to affected individuals by advancing the related discussion.

\paragraph{On the disagreement problem as a power asymmetry.}
The disagreement problem---where different post-hoc XAI methods yield conflicting explanations of the same decision (Sections~\ref{sec:xai-limitations} and ~\ref{sec:agenda-phase4})---is not merely a technical inconvenience.
It creates a structural incentive for deployers to present whichever explanation best supports the decision rather than best enables the affected individual to contest it.
Identifying this as a legal problem under Art.~12(1) GDPR is a deliberate normative choice, intended to foreground this asymmetry.

\subsection*{Researcher Positionality Statement}
\label{ssec:researcher-positionality}

This research was conducted by a team combining EU law expertise with technical XAI research.
While this dual expertise enables us to identify disciplinary misalignments, it also shapes the questions we foreground and the solutions we consider viable.
As such, we privilege formal legal instruments and scientific XAI literature, potentially underrepresenting lived experiences of affected individuals, civil society perspectives, and industry implementation realities---perspectives necessary to make the best of the Right to Explanation.

The researchers involved approach the Right to Explanation as a genuine legal entitlement worth protecting for the individuals it serves.
But the current formulation of the right, combined with the technical capabilities available, creates a significant tension: Some systems might be possible to implement on their own, but cannot be explained sufficiently to fulfill the legal requirements.
Often, a tradeoff between innovation and regulation is stated, but the involved authors believe that unclear regulation is the biggest threat to innovation, which ideally could benefit society at large.
As such, this paper highlights where more work is necessary to provide clear guidance to practitioners on which systems should be developed and how.

\subsection*{Adverse Impact Statement}
\label{sec:adverse-impacts}

\paragraph{Risk of non-deployment of beneficial AI.}
Phase~4 of the four-phase blueprint requires explicit acknowledgment that where no compliant explanation exists, the system must not be deployed.
Applied rigorously, this implies that opaque foundation models may be legally impermissible in the high-risk domains covered by Art.~86 AIA---including healthcare, education, and emergency services (RQ~5).
We believe this is the correct legal interpretation, but acknowledge it may restrict access to AI applications that provide genuine benefits in contexts where interpretable alternatives do not yet achieve comparable performance.
This tension is a policy question that must be addressed transparently in standardization and regulatory work rather than assumed away.
And this should be clarified as early as possible, as developers might shy away from developing such systems in the case of legal uncertainty.

\paragraph{Compliance theater.}
Frameworks such as the Addressee/Purpose Framework (Section~\ref{sec:framework}) are susceptible to performative adoption: deployers could structure explanations to visibly map onto these provisions while still failing to enable meaningful contestation.
The gap between regulatory audit and actual adequacy for affected individuals is a known failure mode of compliance-oriented frameworks, and ours is not immune.

\paragraph{Uneven compliance burden.}
The steps sketched out in the four-phase blueprint require sustained legal and technical expertise to implement.
Large, well-resourced deployers are better positioned to execute this process than public-sector bodies, SMEs, or smaller healthcare and educational institutions---precisely the organizations whose deployments may most directly affect vulnerable populations.
Our framework---as well as other compliance references---may inadvertently advantage incumbents and create barriers for actors with fewer dedicated resources.

\paragraph{Asymmetric reach.}
Ideally, this paper reaches researchers, practitioners, and standardization bodies.
The individuals most directly affected by the explanation gap---those who receive unexplained adverse automated decisions---will not encounter it directly, and may often not be aware of their digital rights.
While we aim to guide the discussion of the Right to Explanation, the pathway from these findings to actual protection depends on uptake by other scholars, whose work may then influence supervisory authorities, courts, and civil society.
We note this as a structural feature of academic impact in rights-protection contexts.

\bibliography{sources.bib}

\clearpage
\appendix

\section{Extended Method Description}
\label{app:method}

A full reporting of the different screening steps and corresponding article numbers is provided in the PRISMA-diagram in Figure~\ref{fig:prisma}.

\subsection*{Search Strings and Databases}

Within both searched databases, Web of Science (WOS) and Scopus, case-insensitive
search strings for title, abstract and keywords of all papers
are used (in WOS called ``Topic'' or TS). The final searches
were conducted on March 23, 2026. The search strings
were---based on the database syntax either with TS for WOS or
TITLE-ABS-KEY for Scopus:

\begin{verbatim}
(XAI OR interpret* OR explain* OR
transparen* OR explanation)
AND ("European Legislation" OR GDPR
OR "General Data Protection Regulation"
OR "Artificial Intelligence Act"
OR "AI Act" OR "European Law")
\end{verbatim}

The first part broadly includes all papers mentioning techniques related to explanation or interpretation of AI systems; the second part requires explicit mention of EU legislation.
The first part may exclude papers arguing, in different terms, that the Right to Explanation does not give rise to technical explanation requirements.
Our review was not designed to assess whether the scholarly debate favors explainability techniques or accepts AI systems with only general explanations; we take the position that a Right to Explanation requires such methods (see Section~\ref{sec:right-to-explanation}).

\subsection*{Inclusion and Exclusion Criteria}

To limit the initial record set, papers in languages other than English and publications before 2024 were excluded.
The temporal restriction to post-2024 literature is motivated by several considerations.

Most directly, Art. 86 AIA---a central provision of this review---was inserted into the final AIA text only shortly before its publication in the Official Journal of the European Union on July 12, 2024; substantive engagement with this provision is therefore structurally confined to literature published thereafter.
Arguably, since the Dun \& Bradstreet judgment clarified the Right to Explanation significantly, papers published between the final version of the AIA and this judgment also lack important information regarding the provisions of the GDPR regarding this right.

Second, the pre-2024 debate---centered on whether the GDPR establishes a Right to Explanation at all, and on early implementation proposals such as counterfactual explanations \cite{Wachter2018}---has been extensively documented in prior works.
Crucially, this earlier scholarship does not drop out of the analysis: The included papers themselves cite and build upon it, allowing us to assess how foundational contributions have been received, extended, or corrected in light of the changed legal framework.
Pre-2024 literature thus enters the review implicitly through the citation practices of the corpus.

Third, the research question of this review---how XAI can satisfy the Right to Explanation under the current EU regulatory framework---is inherently a post-2024 question, as that framework only assumed its present shape with the AIA entering into force. Restricting the corpus to this period therefore aligns scope with research question.


Overall, the search with the criteria as defined before yielded 1827 results in WOS and 1478 in Scopus; after deduplication in Citavi this resulted in 2684 records, of which 41 patents were excluded for their application-specific nature, leaving 2643 records for screening.
Full inclusion and exclusion criteria are listed in Table~\ref{tab:criteria}.

\subsection*{Title/Abstract Screening}

Title/abstract screening was conducted with ASReview~\cite{vandeSchoot2021} by one reviewer with a technical background in XAI, but previous experience in working with legal texts.
ASReview presents an ordered queue of articles following an initial set of random samples.
A stopping criterion of 40 consecutive irrelevant articles was applied, reached after screening 219 title/abstract pairs (roughly 8\% of the dataset).
The authors of ASReview note that ``95\% of the eligible studies will be found after screening between only 8\% to 33\% of the studies''~\cite{vandeSchoot2021}, consistent with our result.
Since our search is focused on a rather specific topic---the implementation of the Right to Explanation via XAI---but with a broad initial search, a comparatively low number of relevant articles seems justified.
Screening yielded 60 articles for full text retrieval.

\subsection*{Validation}

To validate the screening process, an additional random sample of 60 title/abstract pairs from previously unscreened papers was reviewed.
As none met the inclusion criteria, this confirmed prior screening decisions.

\subsection*{Full Text Retrieval and Screening}

Full texts were sought via online sources; for six papers unavailable online, authors were contacted directly.
Five additional full texts were retrieved within a one-week response period; two further texts were excluded for not being available in English, leaving 57 papers for full text assessment.

Two authors---one with a technical background, one with a background in law---screened all full texts against the predefined exclusion criteria in Table~\ref{tab:criteria}.
Uncertain cases were resolved jointly.
Of 57 assessed papers, 19 were included; 23 were excluded for insufficient legal grounding (18 with limited AIA/GDPR coverage, 5 based on---for our research question---irrelevant or outdated laws) and 15 for insufficient XAI content.

During dual full text review, each included paper was coded along four dimensions: (i) XAI classification scheme used; (ii) legal basis(es) cited for the Right to Explanation; (iii) treatment of explanation form; (iv) treatment of explanation purpose/content.
Themes in Section~\ref{sec:findings} correspond to these coding dimensions; the recurrent conflation of (iii) and (iv) across the corpus is what motivated the Addressee/Purpose Framework.


\clearpage

\section{Explanation Properties Used Throughout the Surveyed Literature}
\label{app:properties}

As described in Section~\ref{sec:xai-classification}, the surveyed literature used
different properties to describe what could be desiderata for explanations or for
explainability methods. The most specific approaches were based on two different lists of explainability properties, which are presented in Table~\ref{tab:fresz-properties} and Table~\ref{tab:colmenarejo-properties}.
\citet{Nauta2022}, the basis for the properties by \citet{Fresz2024}, especially mention that of their Co-12-properties, the first six (Correctness, Completeness, Consistency, Continuity, Contrastivity, Covariate Complexity) are content-properties, the next three are presentation-properties (Compactness, Compositionality, Confidence), and the last three are user-properties (Context, Coherence, Controllability).
When comparing these two lists, one can note a few differences, as both lists contain properties the other one either lacks or does not state clearly.
With the formulation of complexity, alternatively called comprehensibility or interpretability, by \citet{Colmenarejo2025}, multiple properties of \citet{Fresz2024} could be meant, namely Covariate Complexity, Contrastivity, Compactness and others.
While this makes the Co-12-properties seem like the more comprehensive list, it also lacks some information such as the subgroup/fairness behavior described as Homogeneity in Table~\ref{tab:colmenarejo-properties}.
Both works state that some tradeoffs between these properties arise in practical implementation.
\citet{Fresz2024} additionally note that some of these properties might compensate for others, e.g.,\ Consilience (using multiple explanation methods) and Confidence (uncertainty estimates of the decision and explanation) in cases where sufficient Correctness cannot be achieved.
The different notions of what an explanation or an explanation method could entail additionally complicate the search for a legally compliant implementation of the Right to Explanation.
To show how the Addressee/Purpose Framework (Section~\ref{sec:framework}) can help here, we provide an exemplary mapping of the properties by \citet{Nauta2022} onto our framework.

\subsection*{Relation to the Addressee/Purpose Framework.}
For the property clusters by \citet{Nauta2022}, the \textbf{Presentation} cluster (Compactness, Compositionality, Confidence) and the \textbf{User} cluster (Context, Coherence, Controllability) correspond to the \emph{form} dimension governed by Art.~12(1) GDPR: What counts as compact, coherent, or contextually appropriate is inherently addressee-dependent.
The \textbf{Content} cluster (Correctness, Completeness, Consistency, Continuity, Contrastivity, Covariate Complexity) corresponds to the \emph{content} dimension governed by the substantive provisions---Art.~15(1)(h) GDPR, Art.~86 AIA, and Art.~18(8)(a) CCD---with Contrastivity being the property most directly mandated by the contestation purpose, and Consistency and Continuity bearing on the disagreement problem (RQ~4).
Two cross-cutting tensions qualify this mapping: Covariate Complexity carries form implications because high feature-interaction complexity reduces intelligibility for non-expert addressees; and Confidence can constitute a content requirement where decision uncertainty is material to contestation.
The Addressee/Purpose Framework thereby imposes a purpose-specific legal ordering on these clusters that Tables~\ref{tab:fresz-properties} and~\ref{tab:colmenarejo-properties} alone do not provide.

\begin{table*}[ht]
\centering
\caption{Co-12-properties used by \citet{Fresz2024}.
The first twelve were defined by \citet{Nauta2022}, whereas the last five were defined by \citet{Fresz2024} as so-called process properties, as they pertain to the process of using XAI. Table by \citet{Fresz2024} and based on \cite{Nauta2022}.}
\label{tab:fresz-properties}
\small
\begin{tabular}{p{4cm}p{8cm}}
\hline
\textbf{Property} & \textbf{Definition} \\
\hline
Correctness &
  Describes how faithful the explanation is w.r.t.\ the black box. \\
\hline
Completeness &
  Describes how much of the black box behavior is described in the explanation. \\
\hline
Consistency &
  Describes how deterministic and implementation-invariant the explanation method is. \\
\hline
Continuity &
  Describes how continuous and generalizable the explanation function is. \\
\hline
Contrastivity &
  Describes how discriminative the explanation is w.r.t.\ other events or targets. \\
\hline
Covariate Complexity &
  Describes how complex the (interactions of) features in the explanation are. \\
\hline
Compactness &
  Describes the size of the explanation. \\
\hline
Compositionality &
  Describes the format and organization of the explanation. \\
\hline
Confidence &
  Describes the presence and accuracy of probability information in the explanation. \\
\hline
Context &
  Describes how relevant the explanation is to the user and their needs. \\
\hline
Coherence &
  Describes how accordant the explanation is with prior knowledge and beliefs. \\
\hline
Controllability &
  Describes how interactive or controllable an explanation is for a user. \\
\hline
\multicolumn{2}{c}{\rule[0.5ex]{2.8cm}{0.4pt}\quad\rule[0.5ex]{5.6cm}{0.4pt}} \\
\hline
Consilience &
  Describes whether more than one XAI method should be used to fulfill the legal
  requirements. \\
\hline
Computations &
  Describes when an explanation needs to be available, e.g.\ right when a prediction is
  shown or at a later point of time. \\
\hline
Coverage &
  Describes what should be explained: a single prediction (local), an entire model
  (global) or the influence of training data samples (for predictions or the overall model
  behavior). \\
\hline
Counterability &
  Describes whether a process needs to be implemented which allows end-users to object to
  AI decisions or explanations. \\
\hline
Constancy &
  Describes whether explanations need to be presented in a format that can be saved for
  later examination. \\
\hline
\end{tabular}
\end{table*}

\begin{table*}[ht]
\centering
\caption{Properties used by \citet{Colmenarejo2025}, based on
their paper.}
\label{tab:colmenarejo-properties}
\small
\begin{tabular}{p{4cm}p{8cm}}
\hline
\textbf{Property} & \textbf{Definition} \\
\hline
Complexity (or comprehensibility/interpretability) &
  How understandable is the explanation to the end-user? Measured e.g.\ via the number of
  premises in an explanatory rule. \\
\hline
Fidelity (or faithfulness) &
  How well does the explanation approximate the machine learning model? Measured e.g.\ as
  the ratio of correct predictions by an explanation over all predictions. \\
\hline
Accuracy &
  How well does the explanation work for a novel data point? Measured e.g.\ as the ratio
  of correct predictions of the explanation compared to the ground truth labels. Accuracy
  may also describe out-of-distribution behavior and the use of probability values in the
  explanation. \\
\hline
Robustness (or sensitivity, stability) &
  How similar are explanations for two different data points? This depends on a formalized
  notion of similarity. \\
\hline
Homogeneity &
  How does faithfulness change across different (sub-)groups? This question is especially
  related to the fairness behavior of the explanation method itself. \\
\hline
\end{tabular}
\end{table*}

\end{document}